\documentclass[a4paper,fleqn,numbers]{cas-dc}
\usepackage{algorithm}
\usepackage{amsmath}
\usepackage{amssymb}
\usepackage{xcolor}
\usepackage{algpseudocode}
\usepackage{graphicx}
\usepackage{caption}  % 可选
\usepackage{makecell}
\usepackage[authoryear,longnamesfirst]{natbib}

\def\tsc#1{\csdef{#1}{\textsc{\lowercase{#1}}\xspace}}
\tsc{WGM}
\tsc{QE}
\begin{document}
\let\WriteBookmarks\relax
\def\floatpagepagefraction{1}
\def\textpagefraction{.001}

% Short title
%\shorttitle{}    

% Short author
%\shortauthors{}  

% Main title of the paper
\title[mode = title]{DAMS:Dual-Branch Adaptive Multiscale Spatiotemporal Framework for Video Anomaly Detection}  

% Title footnote mark
% eg: \tnotemark[1]
%\tnotemark[1] 

% Title footnote 1.
% eg: \tnotetext[1]{Title footnote text}
%\tnotetext[1]{} 

% First author
%
% Options: Use if required
% eg: \author[1,3]{Author Name}[type=editor,
%       style=chinese,
%       auid=000,
%       bioid=1,
%       prefix=Sir,
%       orcid=0000-0000-0000-0000,
%       facebook=<facebook id>,
%       twitter=<twitter id>,
%       linkedin=<linkedin id>,
%       gplus=<gplus id>]

\author[1]{Dezhi An}[orcid=0000-0001-7963-6308]

% Corresponding author indication
\cormark[1]

% Footnote of the first author
\fnmark[1]

% Email id of the first author
\ead{andezhi@gsupl.edu.cn}

% URL of the first author
%\ead[url]{...}
% Credit authorship
% eg: \credit{Conceptualization of this study, Methodology, Software}
\credit{Conceptualization}

\affiliation[1]{organization={School of Cyberspace Security,Gansu University of Political Science and Law},
            addressline={No. 6 Anning West Road}, 
            city={Lanzhou},
            postcode={730070}, 
            state={Gansu},
            country={China}}

\author[2]{Wenqiang Liu}[orcid=0009-0000-9979-2164]
% Footnote of the second author
\fnmark[2]

% Email id of the second author
\ead{liuwenqiang@stu.gsupl.edu.cn}

% URL of the second author
%\ead[url]{}

% Credit authorship
\credit{Software,Writing-original draft}

\affiliation[2]{organization={School of Cyberspace Security,Gansu University of Political Science and Law},
            addressline={No. 6 Anning West Road}, 
            city={Lanzhou},
            postcode={730070}, 
            state={Gansu},
            country={China}}

\author[3]{Kefan Wang}%[]
% Footnote of the second author
\fnmark[3]

% Email id of the second author
\ead{wkf6856@163.com}

% URL of the second author
%\ead[url]{}

% Credit authorship
\credit{Validation}

\affiliation[3]{organization={School of Cyberspace Security,Gansu University of Political Science and Law},
            addressline={No. 6 Anning West Road}, 
            city={Lanzhou},
            postcode={730070}, 
            state={Gansu},
            country={China}}

\author[4]{Zening chen}%[]
% Footnote of the second author
\fnmark[4]

% Email id of the second author
\ead{czn3361@163.com}

% URL of the second author
%\ead[url]{}

% Credit authorship
\credit{Investigation}

\affiliation[4]{organization={School of Cyberspace Security,Gansu University of Political Science and Law},
            addressline={No. 6 Anning West Road}, 
            city={Lanzhou},
            postcode={730070}, 
            state={Gansu},
            country={China}}

\author[5]{Jun Lu}[orcid=0009-0009-9387-6733]
% Footnote of the second author
\fnmark[5]

% Email id of the second author
\ead{lj6703@gsupl.edu.cn}

% URL of the second author
%\ead[url]{}

% Credit authorship
\credit{Writing-review & editing}

\affiliation[5]{organization={School of Cyberspace Security,Gansu University of Political Science and Law},
            addressline={No. 6 Anning West Road}, 
            city={Lanzhou},
            postcode={730070}, 
            state={Gansu},
            country={China}}

\author[6]{Shengcai Zhang}[orcid=0009-0001-9322-0672]
% Footnote of the second author
\fnmark[6]

% Email id of the second author
\ead{zsc6731@gsupl.edu.cn}

% URL of the second author
%\ead[url]{}

% Credit authorship
\credit{Supervision}

\affiliation[6]{organization={School of Cyberspace Security,Gansu University of Political Science and Law},
            addressline={No. 6 Anning West Road}, 
            city={Lanzhou},
            postcode={730070}, 
            state={Gansu},
            country={China}}

% Corresponding author text
\cortext[1]{Corresponding author}
% Footnote text
%\fntext[1]{}

% For a title note without a number/mark
%\nonumnote{}

% Here goes the abstract
\begin{abstract}
The goal of video anomaly detection is tantamount to performing spatio-temporal localization of abnormal events in the video. The multiscale temporal dependencies, visual-semantic heterogeneity, and the scarcity of labeled data exhibited by video anomalies collectively present a challenging research problem in computer vision. This study offers a dual-path architecture called the Dual-Branch Adaptive Multiscale Spatiotemporal Framework (DAMS), which is based on multilevel feature decoupling and fusion, enabling efficient anomaly detection modeling by integrating hierarchical feature learning and complementary information. The main processing path of this framework integrates the Adaptive Multiscale Time Pyramid Network (AMTPN) with the Convolutional Block Attention Mechanism (CBAM). AMTPN enables multigrained representation and dynamically weighted reconstruction of temporal features through a three-level cascade structure (time pyramid pooling, adaptive feature fusion, and temporal context enhancement). CBAM maximizes the entropy distribution of feature channels and spatial dimensions through dual attention mapping. Simultaneously, the parallel path driven by CLIP introduces a contrastive language-visual pre-training paradigm. Cross-modal semantic alignment and a multiscale instance selection mechanism provide high-order semantic guidance for spatio-temporal features. This creates a complete inference chain from the underlying spatio-temporal features to high-level semantic concepts. The orthogonal complementarity of the two paths and the information fusion mechanism jointly construct a comprehensive representation and identification capability for anomalous events. Extensive experimental results on the UCF-Crime and XD-Violence benchmarks establish the effectiveness of the DAMS framework. \nocite{*}%% Remove this line from your manuscript.
\end{abstract}

% Use if graphical abstract is present
%\begin{graphicalabstract}
%\includegraphics{}
%\end{graphicalabstract}

% Research highlights
%\begin{highlights}
%\item 
%\item 
%\item 
%\end{highlights}

%\nocite{*}

% Keywords
% Each keyword is seperated by \sep

\maketitle

\section{Introduction}

Video anomaly detection (VAD) plays a vital role in intelligent surveillance, public safety, and behavior monitoring~\cite{delic2025sequential,zhao2025smarthome,huang2025track,li2025memoryout,zhu2025vau,huang2025vad}. However, real-world anomalous events are inherently sparse, diverse, and temporally complex, making large-scale frame-level annotation costly and impractical. As a remedy, weakly supervised video anomaly detection (WS-VAD) has garnered attention by relying solely on video-level labels. While efficient, this setting presents key challenges: the open-set nature of anomalies, multi-scale temporal irregularity, and the limited granularity of supervision.

\begin{figure}[t]
    \centering
    \includegraphics[width=\linewidth]{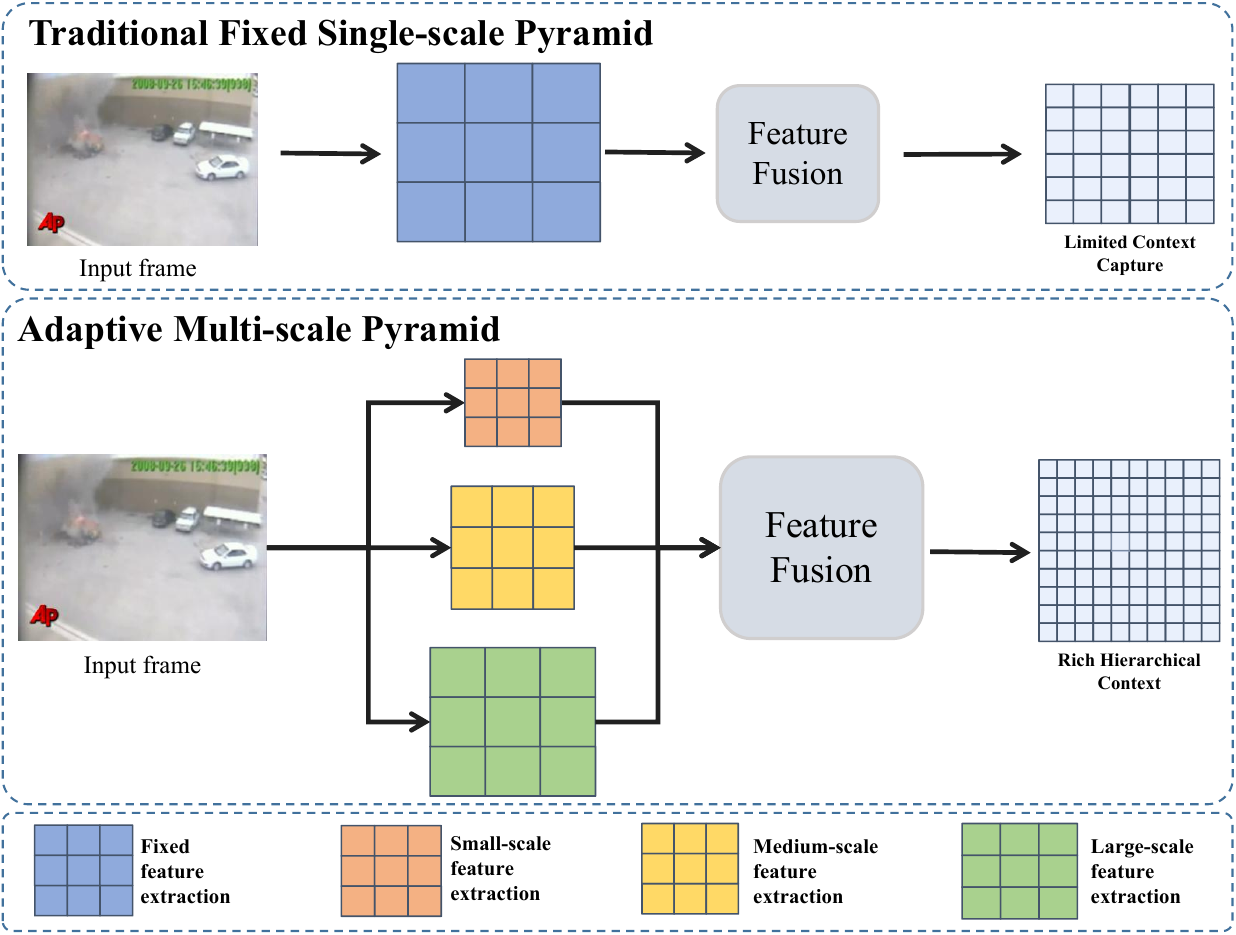}
    \vspace{-0.2cm}
    \caption{{\fontfamily{ptm}\selectfont Comparison between (top) the traditional fixed single-scale pyramid architecture and (bottom) the proposed adaptive multi-scale pyramid architecture. The former processes input frames through a single path, resulting in limited context-capturing capability. In contrast, the latter enhances the representation by processing features at multiple scales in parallel, thereby capturing spatiotemporal information of varying granularities more effectively.}}
    \label{fig1.pdf}
    \vspace{-0.4cm}
\end{figure}

As shown in Figure~\ref{fig1.pdf}, most existing WS-VAD methods adopt single-path architectures and fixed-scale temporal pyramids, which restrict their capacity to capture hierarchical semantics and fine-grained spatiotemporal cues. To address this, we propose a novel dual-path framework that jointly enhances temporal dependency modeling and cross-dimensional feature representation. Our method, DAMS (Dual-path Adaptive Multiscale Spatiotemporal framework), introduces a cascaded Adaptive Multiscale Temporal Pyramid Network (AMTPN) for dynamic temporal modeling, a Convolutional Block Attention Module (CBAM) for channel-spatial calibration, and a CLIP-driven semantic enhancement path for high-level conceptual guidance. This architecture facilitates bidirectional information flow between low-level dynamics and high-level semantics, advancing the modeling of complex, heterogeneous anomalies under weak supervision.

Previous WS-VAD pipelines~\cite{fan2024weakly,feng2021mist,li2022self,tian2021weakly,wu2022self,zaheer2020claws} typically follow a two-stage design: frame-level feature extraction via C3D~\cite{tran2015learning,sultani2018real} or I3D~\cite{carreira2017quo}, followed by multi-instance learning (MIL)~\cite{andrews2002support,li2015multiple} for anomaly localization. While some methods incorporate graph-based~\cite{defferrard2016convolutional} or attention-based~\cite{vaswani2017attention} temporal modeling, they face trade-offs between contextual awareness and noise sensitivity. Notably, prior works~\cite{wu2020not,tian2021weakly} attempt to fuse local and global contexts in parallel structures but suffer from increased computational cost and limited scalability. Furthermore, current methods rely on static feature extraction pipelines and predefined temporal resolutions, limiting adaptability to anomalies of varying durations and semantics.

Recent advances in multimodal fusion~\cite{sun2024multimodal,meng2025audio,wu2024vadclip} explore collaborative learning across RGB, optical flow, and audio modalities. While these approaches offer complementary perspectives, most fusion strategies adopt shallow operations such as concatenation or averaging, lacking hierarchical semantic alignment and adaptive selection mechanisms. Moreover, Transformer-based architectures~\cite{ding2024learnable,ghadiya2024cross,zhou2023dual} emphasize spatial structure while underutilizing temporal dynamics. Collectively, these limitations highlight a fundamental bottleneck: the inability of single-path frameworks to jointly encode fine-grained spatiotemporal structure and abstract semantic understanding, thereby limiting generalization, robustness, and interpretability.

To overcome these challenges, we introduce DAMS, a dual-branch framework built upon hierarchical feature decoupling and cross-path interaction. The main temporal path integrates AMTPN and CBAM: AMTPN employs cascaded temporal pyramid pooling, dynamic weighting, and context-aware recalibration to learn adaptive multiscale temporal representations; CBAM refines channel-spatial features through attention. Complementarily, a parallel CLIP-guided semantic path performs contrastive alignment and hierarchical instance refinement, bridging low-level motion with high-level semantics. This design enables comprehensive and interpretable anomaly detection by integrating structural precision with semantic abstraction.
In a nutshell, our main contributions are as follows:
\begin{itemize}
  \item[\textcolor{black}{$\bullet$}] We propose {DAMS}, a novel dual-path adaptive spatiotemporal framework for weakly supervised video anomaly detection. It jointly models fine-grained temporal dependencies and high-level semantics through hierarchical feature decoupling and bidirectional cross-path fusion.

  \item[\textcolor{black}{$\bullet$}] We design an {Adaptive Multiscale Temporal Pyramid Network (AMTPN)} that captures multiscale temporal patterns via cascaded temporal pooling, content-adaptive fusion, and context-aware enhancement. This structure overcomes the rigidity of fixed temporal resolutions in existing methods.

  \item[\textcolor{black}{$\bullet$}] We incorporate {CBAM-based attention optimization} to recalibrate representations across channel and spatial dimensions, promoting discriminative feature learning through cross-dimensional interaction and information entropy maximization.

  \item[\textcolor{black}{$\bullet$}] We introduce a {CLIP-guided semantic enhancement path} that leverages contrastive vision-language pretraining to guide anomaly localization. It enables adaptive instance selection and semantic alignment, bridging low-level visual patterns and high-level conceptual understanding.

  \item[\textcolor{black}{$\bullet$}] Extensive experiments on multiple VAD benchmarks demonstrate that DAMS achieves superior performance and interpretability compared to state-of-the-art baselines, validating its effectiveness in modeling complex and heterogeneous anomalies.
\end{itemize}

\section{Related Work}
\subsection{Video Anomaly Detection}  %和任务相关
Video anomaly detection based on deep learning continues to be extensively studied through various feature representation methodologies. Early methods primarily rely on reconstruction-based frameworks, where the work in~\cite{hasan2016learning} implements a convolutional autoencoder to reconstruct normal events and detect anomalies through reconstruction errors. Subsequently, the TSC~\cite{luo2017revisit} method employs time-coherent sparse encoding to capture the regularity of normal patterns. Researchers propose a prediction model ~\cite{liu2018future} for forecasting future frames, which leverages the difference between a predicted future frame and its ground truth to detect an abnormal event. However, these approaches mainly employ fixed-scale temporal convolution or recursive architectures, thus limiting their ability to capture inherent multi-scale temporal dependencies essential for characterizing anomalies. Furthermore, as evidenced by the CLAWS~\cite{zaheer2020claws} and MNAD~\cite{park2020learning} methods, traditional feature extraction pipelines generally adhere to static paradigms, exhibiting limited adaptability to dynamic temporal context and semantic significance, significantly compromising their discriminative capacity in complex non-static video sequences.

These limitations stimulate extensive research on attention-enhanced representations. A spatiotemporal attention mechanism~\cite{wang2017spatiotemporal} is proposed, which models long-range dependencies in video sequences. Similarly, MNAD~\cite{park2020learning} proposes a framework with memory-guided normality for anomaly detection. A graph convolutional neural network~\cite{zhong2019graph} is also formulated to capture the topological relationships among video clips. However, these methods predominantly focus on a single feature dimension, thus failing to integrate cross-dimensional feature relationships systematically. Notably, the convolutional block attention module (CBAM)~\cite{woo2018cbam} is proposed for general image recognition tasks, demonstrating the effectiveness of sequential channel and spatial attention mechanisms. However, CBAM's potential for video anomaly detection remains relatively unexplored, especially in combination with multiscale temporal modeling. To address these limitations, the Adaptive Multiscale Temporal Pyramid Network (AMTPN) is introduced, which systematically integrates multiscale temporal modeling with cross-dimensional attention optimization, establishing a novel framework for dynamic and context-aware feature representations in video anomaly detection.

\subsection{Spatiotemporal Modeling Architectures}  % 和你的AB模块技术相关
Recent advances in video anomaly detection have gradually emphasized the importance of effective spatiotemporal modeling. The hierarchical fusion strategy~\cite{wang2017spatiotemporal} integrates spatial and temporal features into a pyramid structure, facilitating mutual enhancement. This approach introduces the space-time compact bilinear operator into video analysis, effectively capturing comprehensive interactions between spatial and temporal features. MSM-Net~\cite{cai2021video} employs a temporal memory network to capture abnormal patterns across different temporal scales. Similarly, a space-time decoupled dual-branch network~\cite{yang2025video} is proposed, wherein spatial and temporal branches independently address specific proxy tasks. To enhance the interpretability of abnormal events, the STPrompt~\cite{wu2024weakly} method incorporates a spatial attention aggregation mechanism that filters irrelevant backgrounds for temporal anomaly detection.Additionally, this method integrates a large language model and a training-free anomaly localization approach, enabling spatial capture through fine-grained text prompts. Further progress in this field emerges from the spatiotemporal pseudo-anomaly generator~\cite{rai2024video} approach. This approach presents a unified video anomaly detection framework based on three distinct anomaly behavior indicators: reconstruction quality, temporal irregularity, and semantic inconsistency.

Despite these notable contributions, existing methods typically employ either predetermined temporal scale selections or static attention distributions, failing to adapt to the inherently variable nature of anomalous events. The present study proposes a novel AMTPN-CBAM architecture that synergistically integrates adaptive multi-scale temporal pyramid networks with convolutional block attention modules to address these limitations. Unlike previous approaches that treat temporal scale selection and attentional processing as disconnected components, the proposed framework implements a unified mechanism that modulates temporal receptive fields through content-aware attention signals. The AMTPN component constructs a hierarchical pyramid of temporal representations with learnable scale adaptation parameters. In contrast, the integrated CBAM refines these features through channel and spatial attention mechanisms specifically calibrated for anomaly detection. This integration facilitates more effective modelling of complex temporal dynamics characterizing anomalous events, particularly those with inconsistent durations and irregular motion patterns, thereby significantly enhancing state-of-the-art video anomaly detection performance.
\subsection{Dual-path architecture paradigm}
Multimodal approaches and diffusion models for video anomaly detection demonstrate significant potential by utilizing complementary information sources. Audiovisual fusion approaches~\cite{tian2018audio} pioneered research on event detection, while VadCLIP~\cite{wu2024vadclip} leveraged the fine-grained correlation between vision and language through a dual-branch architecture. One branch utilizes visual features for coarse-grained binary classification in this architecture, while the other exploits fine-grained language-image alignment. Furthermore, AR-Net~\cite{wan2021weakly} introduces multi-instance learning to leverage video-level supervision. DCMD~\cite{wang2025dual} proposes a deep diffusion model framework incorporating content and motion conditioning. Despite these advances, existing multimodal fusion strategies and diffusion model frameworks predominantly employ basic cascading or weighted averaging operations, lacking the capability for hierarchical semantic decoupling and adaptive recombination of heterogeneous information sources. Furthermore, these methods typically implement single-path architectures that process different modalities in parallel before merging, failing to establish bidirectional interactions between complementary feature hierarchies.

Recent advances in large-scale pre-trained models have enabled significant progress in knowledge transfer for video understanding. The introduction of Contrastive Language-Image Pre-training (CLIP)~\cite{radford2021learning} has propelled computer vision into a new era, demonstrating excellent zero-shot learning transfer capabilities in various visual recognition tasks through joint vision-language embedding. Building upon this foundation, MMVAD~\cite{biswas2025mmvad} and VadCLIP~\cite{wu2024vadclip} investigated the potential of CLIP in video anomaly detection by exploring prompt learning and vision-language alignment, respectively. However, these methods primarily utilize CLIP as a standalone feature extractor or classifier, failing to systematically integrate high-level semantic understanding with complementary low-level spatiotemporal processing. Furthermore, these methods predominantly employ unified feature extraction, neglecting the fundamental multi-scale temporal dependencies essential for characterizing anomalies. To address these limitations, this study proposes a principled dual-path architecture that facilitates bidirectional information exchange between bottom-up spatiotemporal processing via AMTPN-CBAM and top-down semantic guidance via CLIP. This novel framework enables seamless integration of multi-scale temporal modeling with cross-modal semantic alignment, facilitating comprehensive anomaly representation and interpretation that transcends the capabilities of existing single-path methods.

\section{Proposed Method}
\begin{figure*}[t]
    \centering
    \includegraphics[width=\linewidth]{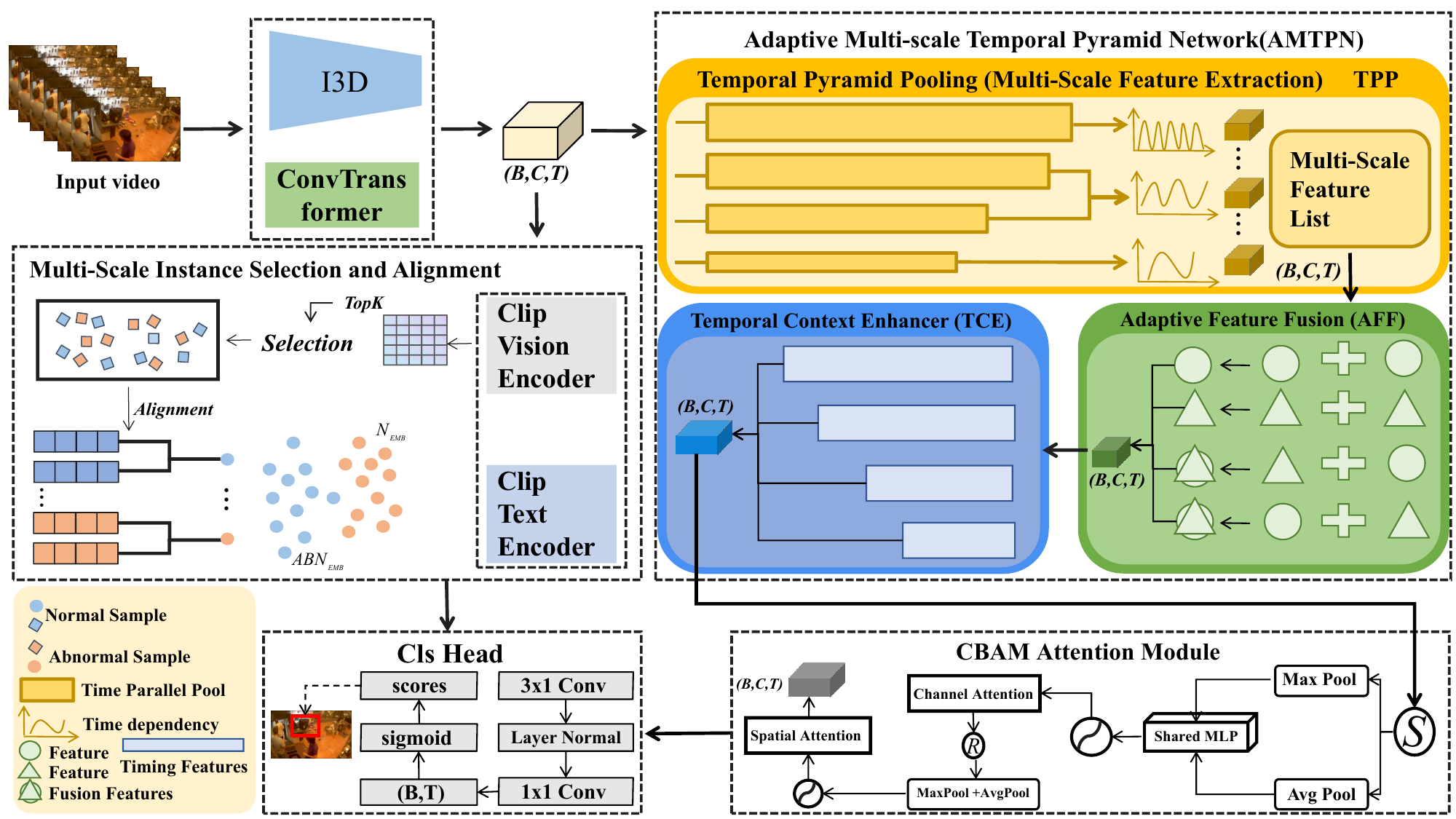}
    \caption{{\fontfamily{ptm}\selectfont The comprehensive architecture of our proposed video anomaly detection framework. The system consists of multiple key components: (1) Feature extraction backbone utilizing either I3D and ConvTransformer for initial video representation; (2) Adaptive Multi-scale Temporal Pyramid Network (AMTPN) comprising three critical modules: Temporal Pyramid Pooling (TPP) for multi-scale feature extraction, Adaptive Feature Fusion (AFF) for intelligent feature integration, and Temporal Context Enhancer (TCE) for contextual information enrichment; (3) Multi-Scale Instance Selection and Alignment mechanism leveraging CLIP vision and text encoders for semantic guidance; (4) CBAM Attention Module with parallel channel and spatial attention pathways for refined feature representation; and (5) Classification Head for anomaly detection.}}
    \label{fig:fig2}
    \vspace{-0.2cm}
\end{figure*}

\subsection{Overview}
The proposed architecture systematically integrates adaptive spatio-temporal feature learning and cross-modal semantic representation within an information-theoretic framework. As illustrated in Figure \ref{fig:fig2}, the methodological framework comprises two complementary and parallel pathways: the spatio-temporal feature representation branch and the visual-linguistic semantic alignment branch. The spatiotemporal representation branch implements an Adaptive Multi-scale Temporal Pyramid Network (AMTPN), which hierarchically extracts multi-resolution temporal features through differentiable heterogeneous temporal receptive fields, subsequently processing these features via a Convolutional Block Attention Module (CBAM). The CBAM module complements the temporal modelling capability of AMTPN in Euclidean space by sequentially applying channel-wise and spatial attention transformation mechanisms. Consequently, a comprehensive spatio-temporal representation manifold is constructed. These spatio-temporal features are subsequently processed by the parameterized Adaptive Feature Fusion (AFF) module, which employs discriminative metrics based on the entropy minimization principle to optimize the weight distribution of dynamically modulated multi-scale feature representations. In parallel, the semantic guidance branch leverages the Contrastive Language-Image Pre-training (CLIP) paradigm to establish a high-dimensional embedding space, projecting normal and abnormal patterns into distributional representations within shared visual-language manifolds through self-supervised contrastive learning while facilitating cross-modal knowledge transfer through multi-scale instance selection and probabilistic alignment. The integration of the two pathways is achieved through an attention-driven mechanism based on mutual information maximization, enabling the framework to comprehensively capture the complex spatio-temporal dynamic characteristics of anomalous events across multi-scale and multi-semantic levels. 

\subsection{Adaptive Multi-Scale Time Pyramid Network}

\begin{algorithm}[htbp]
\caption{Adaptive Multi-scale Temporal Pyramid Network (AMTPN)}
\label{alg:amtpn}
\textbf{Require:} video feature sequence $X \in \mathbb{R}^{B \times C \times T}$, temporal scales set $S = \{s_1, s_2, ..., s_k\}$ \\
\textbf{Ensure:} enhanced temporal feature representation $F_{\text{out}}$
\begin{algorithmic}[1]
\State $F_{\text{multi}} \gets \emptyset$ \Comment{collection of multi-scale features}
\State $W \gets \emptyset$ \Comment{adaptive fusion weights}
\State \textcolor{gray}{// Stage 1: Temporal Pyramid Pooling (TPP)}
\For{each $s \in S$}
    \State $kernel\_size \gets s$
    \State $padding \gets s \mathbin{//} 2$ \Comment{maintain temporal dimension}
    \State $F_{\text{tmp}} \gets \text{TemporalConv1D}(X, kernel\_size, padding)$
    \State $F_{\text{tmp}} \gets \text{MaxPool1D}(F_{\text{tmp}}, kernel\_size, padding)$
    \State $F_{\text{multi}} \gets F_{\text{multi}} \cup \{F_{\text{tmp}}\}$
\EndFor
\State \textcolor{gray}{// Stage 2: Adaptive Feature Fusion (AFF)}
\State $E \gets \emptyset$
\For{each $F_{\text{tmp}} \in F_{\text{multi}}$}
    \State $e \gets \text{GlobalAvgPool1D}(F_{\text{tmp}})$
    \State $e \gets \text{MLP}(e)$ \Comment{2-layer FC with ReLU}
    \State $E \gets E \cup \{e\}$
\EndFor
\State $E_{\text{concat}} \gets \text{Concatenate}(E)$
\State $W \gets \text{Softmax}(\text{FC}(E_{\text{concat}}))$
\State $F_{\text{agg}} \gets 0$
\For{$i = 1$ to $|S|$}
    \State $F_{\text{agg}} \gets F_{\text{agg}} + W[i] \times F_{\text{multi}}[i]$
\EndFor
\State \textcolor{gray}{// Stage 3: Temporal Context Enhancer (TCE)}
\State $Q \gets \text{LinearProj}(F_{\text{agg}})$
\State $K \gets \text{LinearProj}(F_{\text{agg}})$
\State $V \gets \text{LinearProj}(F_{\text{agg}})$
\State $A \gets \text{Softmax}(\text{MatMul}(Q, K^\top) / \sqrt{d_k})$
\State $C \gets \text{MatMul}(A, V)$
\State $F_{\text{ctx}} \gets \text{LayerNorm}(F_{\text{agg}} + C)$
\State $F_{\text{out}} \gets \text{LayerNorm}(F_{\text{ctx}} + \text{FFN}(F_{\text{ctx}}))$
\State \Return $F_{\text{out}}$
\end{algorithmic}
\end{algorithm}

\subsubsection{Temporal Pyramid Pooling (TPP)}
The TPP module systematically generates hierarchical multiscale temporal features by applying adaptive pooling operations across different temporal receptive fields. Given input features $X \in \mathbb{R}^{B \times C \times T}$, the multiscale feature set is defined as:
\begin{equation}
\mathcal{F}_{\text{multi}} = \{\phi_{s_1}(X), \phi_{s_2}(X), ..., \phi_{s_K}(X)\}.
\end{equation}
where $S = \{s_1, s_2, ..., s_K\}$ represents the set of temporal scales . Each scale-specific transformation $\phi_{s_k}(X)$ is implemented as:
\begin{equation}
\phi_{s_k}(X) = \text{ReLU}(\text{BN}(\text{Conv1D}(\mathcal{P}_{s_k}(X)))).
\end{equation}
where $\mathcal{P}_{s_k}(X) = \text{AvgPool1D}(X; s_k, 1, \lfloor s_k/2 \rfloor)$ denotes average pooling with kernel size $s_k$, stride 1, and padding $\lfloor s_k/2 \rfloor$, and $\text{Conv1D}$ represents 1×1 convolution with learnable parameters $\theta_{s_k}$. BN denotes batch normalization, and ReLU is the activation function.This hierarchical pooling strategy enables the capture of temporal patterns at different granularities: fine-grained local patterns at smaller scales and coarse-grained global patterns at larger scales, providing comprehensive temporal context for anomaly detection.

\subsubsection{Adaptive Feature Fusion (AFF)}
The AFF module addresses the challenge of optimally combining multi-scale temporal features with heterogeneous characteristics. Rather than using fixed fusion weights, the AFF module learns adaptive weights that reflect the relative importance of different temporal scales for anomaly detection.
Architecture Design: Given the multi-scale feature set $\mathcal{F}_{\text{multi}} = \{\phi_{s_1}(X), ..., \phi_{s_K}(X)\}$, the AFF module generates adaptive fusion weights through a lightweight attention mechanism:
\begin{equation}
w = \text{Softmax}(\text{Conv1D}(\text{AdaptiveAvgPool1D}(\phi_{s_1}(X)))),
\end{equation}
where $w \in \mathbb{R}^K$ represents the normalized attention weights satisfying $\sum_{i=1}^K w_i = 1$ and $w_i \geq 0$. The adaptive average pooling operation captures global temporal statistics, while the 1D convolution learns scale-specific importance patterns.Fusion Process: The final fused representation is computed as:
\begin{equation}
\mathcal{F}_{\text{fused}} = \text{Conv1D}\left(\sum_{i=1}^K w_i \cdot \phi_{s_i}(X)\right).
\end{equation}
where the additional convolution layer refines the weighted combination to produce the final multi-scale representation.

\subsubsection{Time Context Enhancer (TCE)}
The TCE module employs a channel-wise attention mechanism to enhance temporal feature representations by adaptively weighting channel importance. The core computation is formulated as:
\begin{equation}
\text{TCE}(\mathcal{F}_{\text{agg}}) = \mathcal{F}_{\text{agg}} \odot \sigma(W_2 \cdot \text{ReLU}(W_1 \cdot \text{GAP}(\mathcal{F}_{\text{agg}}))).
\end{equation}
where $\mathcal{F}_{\text{agg}}$ denotes the fused feature output from the AFF module, $\text{GAP}(\cdot)$ represents global average pooling along the temporal dimension, $W_1$ and $W_2$ are learnable projection matrices with reduction ratio $r$, $\sigma(\cdot)$ is the sigmoid activation, and $\odot$ denotes element-wise multiplication.The attention mechanism learns adaptive channel importance weights, effectively constructing data-driven feature recalibration in the channel space. The learned weights $\alpha_c = \sigma(W_2 \cdot \text{ReLU}(W_1 \cdot z_c))$ encode the relevance of each channel $c$ for anomaly detection, where $z_c$ represents the global temporal context for channel $c$.Through this adaptive weighting scheme, the TCE module maximizes the discriminative capacity of informative channels while suppressing less relevant ones, significantly improving the model's ability to distinguish complex temporal anomaly patterns and capture long-term contextual dependencies.

\subsubsection{Information Flow Analysis}
From an information processing perspective, AMTPN implements a hierarchical feature transformation pipeline: 
\begin{equation} 
X \rightarrow \{\phi_s(X)\} \rightarrow \sum_s w_s\phi_s(X) \rightarrow \text{TCE}\left(\sum_s w_s\phi_s(X)\right). 
\end{equation}
As detailed in Algorithm~\ref{alg:amtpn}, the computational pipeline consists of three sequential stages: multi-scale feature extraction (TPP), adaptive fusion (AFF), and contextual enhancement (TCE).
\textbf{Information-Theoretic Framework}: AMTPN preserves task-relevant information while enhancing discriminative capacity. The design objective is: 
\begin{align}
\max_{\theta} I(f_\theta(X); Y) \quad \text{subject to} \quad H(f_\theta(X)|X) \leq \epsilon. 
\end{align} where $f_\theta$ 
represents the AMTPN transformation, and $\epsilon$ bounds information loss. \textbf{Multi-Scale Complementarity Analysis}: To quantify synergistic benefits of multi-scale features, we define a Complementarity Index (CI): 
\begin{align} 
\text{CI}(s_i, s_j) &= \frac{I_{ij} - I_{\max}}{I_i + I_j}. 
\end{align} 
where $I_i = I(\phi_{s_i}(X); Y)$, $I_j = I(\phi_{s_j}(X); Y)$, $I_{ij} = I(\phi_{s_i}(X), \phi_{s_j}(X); Y)$, and $I_{\max} = \max(I_i, I_j)$. This measures relative information gain from combining scales compared to the best single-scale feature.

\subsection{Deep Integration of Convolutional Block Attention Module}
CBAM implements progressive reweighting of feature tensors. Given input feature map $\mathbf{F} \in \mathbb{R}^{B \times C \times T}$, CBAM applies two sequential attention operations:
\begin{align}
\mathbf{F}' &= \mathbf{M}_c(\mathbf{F}) \otimes \mathbf{F}, \\
\mathbf{F}'' &= \mathbf{M}_t(\mathbf{F}') \otimes \mathbf{F}'.
\end{align}
where $\mathbf{M}_c\in\mathbb{R}^{B \times C \times 1}$ and $\mathbf{M}_t\in\mathbb{R}^{B \times 1 \times T}$ represent channel and temporal attention maps, and $\otimes$ denotes element-wise multiplication. Channel attention focuses on "which feature channels are important":
\begin{equation}
\mathbf{M}_c(\mathbf{F})=\sigma(f_c(\text{AvgPool}_T(\mathbf{F}))+f_c(\text{MaxPool}_T(\mathbf{F}))).
\end{equation}
where $f_c$ represents a shared neural network with two fully connected layers, and $\sigma$ denotes sigmoid activation. Temporal attention focuses on "which temporal positions are important":
\begin{equation}
\mathbf{M}_t(\mathbf{F}')=\sigma(f_t([\text{AvgPool}_C(\mathbf{F}'); \text{MaxPool}_C(\mathbf{F}')])).
\end{equation}
where $f_t$ is a convolutional layer with kernel size $k$, and $[\cdot; \cdot]$ denotes channel-wise concatenation. For temporal sequence processing, dimensional transformations are required:
\begin{align}
\mathbf{F}_{2D} &= \text{Unsqueeze}(\mathbf{F}, \text{dim}=2) \in \mathbb{R}^{B \times C \times 1 \times T}, \\
\mathbf{F}''_{1D} &= \text{Squeeze}(\text{CBAM}(\mathbf{F}_{2D}), \text{dim}=2) \in \mathbb{R}^{B \times C \times T}.
\end{align}

From an information theory perspective, CBAM optimizes feature representation by solving a mutual information maximization problem:
\begin{equation}
\max_{\mathbf{M}_c, \mathbf{M}_t} I((\mathbf{M}_c \otimes \mathbf{M}_t \otimes \mathbf{F}), \mathbf{Y}).
\end{equation}

\subsection{Dual-branch Architecture}
We propose an innovative dual-branch collaborative architecture that enables efficient video anomaly detection by integrating a main processing path with a Contrastive Language-Image Pre-training (CLIP) path in parallel. This architecture addresses two core challenges: long-term temporal modelling and insufficient semantic representation.
The main processing path adopts a Video Anomaly Detection Transformer (VADTransformer) structure comprising a temporal convolutional transformer backbone and a feature pyramid neck network incorporating two key components: an Adaptive Multi-scale Temporal Pyramid Network (AMTPN) and a Convolutional Block Attention Module (CBAM). AMTPN enhances perception capabilities for temporal patterns through multi-scale temporal pyramid pooling with adaptive weighted fusion:
\begin{align} F_{\text{temp}} &= \text{AMTPN}(F_{\text{input}}) \notag \\ &= \text{TCE}\left( \text{AFF}\left( \{TPP_s(F_{\text{input}})\}_{s \in \{1,3,9,27\}} \right) \right). \end{align}
where $TPP_s$ denotes temporal pooling at scale $s$, AFF represents adaptive feature fusion, and TCE denotes temporal context enhancement.
In parallel, the CLIP path generates frame-level pseudo-labels by computing similarity between video frame features and textual descriptions of anomaly classes:
\begin{equation} S_{\text{clip}}(I_t, T_c) = \frac{\exp(\text{sim}(E_v(I_t), E_t(T_c))/\tau)}{\sum_{c'}\exp(\text{sim}(E_v(I_t), E_t(T_{c'}))/\tau)}. \end{equation}
where $I_t$ represents the video frame at time $t$, $T_c$ denotes textual description for anomaly class $c$, $E_v$ and $E_t$ are CLIP's visual and textual encoders, $\text{sim}$ is cosine similarity, and $\tau$ is the temperature parameter. These pseudo-labels provide additional supervision signals during training, enhancing semantic understanding of anomalous events.

The dual-branch architecture workflow can be formally represented as a parallel processing system for efficient information integration. The CLIP path functions as an offline prior knowledge provider, generating frame-level pseudo-labels through vision-text matching during preprocessing:
\begin{equation} 
P_{\text{clip}}(y_t|I_t) = \text{softmax}(\lambda \cdot \text{cos}(E_v(I_t), E_t(T_{\text{abn}}))). 
\end{equation}
where $E_v$ and $E_t$ represent CLIP's visual and textual encoders, $\lambda$ denotes a scaling parameter, and $T_{\text{abn}}$ corresponds to textual descriptions for anomaly classes. These pseudo-labels are pre-computed and stored, requiring no additional annotation while leveraging CLIP's generalisation capability.
During training, the main processing path utilises VADTransformer to capture intrinsic spatiotemporal dependencies:
\begin{align} 
F_{\text{vad}} = \Phi(F_{\text{input}}) &= \text{ClsHead}(\text{CBAM}(\text{AMTPN}( \nonumber \\ &\quad F_{\text{backbone}}(F_{\text{input}})))). 
\end{align}

\subsection{Loss Function}  %看页数需要不够就加，页数够了就不要，可选。
Our model employs a multi-task learning framework with three complementary loss components automatically balanced through uncertainty-based weighting. The comprehensive loss function addresses the challenges of weakly supervised video anomaly detection by integrating semantic supervision, video-level constraints, and feature discrimination.

\subsubsection{Pseudo-label Supervised Loss}
The pseudo-label supervised loss leverages CLIP-generated semantic labels to provide fine-grained supervision at the frame level. Given the inherent class imbalance in anomaly detection tasks, we employ focal loss to address the dominance of normal frames:
\begin{align}
\mathcal{L}_{\text{pse}} = -\alpha_t (1-p_t)^\gamma \log(p_t).
\end{align}
where $p_t$ represents the predicted probability for the true class, $\alpha_t$ is the weighting factor for class balance, and $\gamma$ is the focusing parameter that down-weights easy examples. The pseudo-labels are generated by thresholding CLIP similarity scores: $\hat{y}t = \mathbb{I}[P{\text{clip}}(y_t|I_t) > \tau]$, where $\tau$ is the threshold parameter and $\mathbb{I}[\cdot]$ is the indicator function.

\subsubsection{Video-level Classification Loss}
We implement a classification loss that operates on aggregated video representations to ensure robust video-level anomaly detection. The video-level score is computed through top-k average pooling to focus on the most anomalous segments:
\begin{align}
s_{\text{video}} = \frac{1}{k}\sum_{i=1}^{k} s_{(i)},
\end{align}
where $s_{(i)}$ denotes the $i$-th highest frame-level anomaly score and $k = \lceil 0.1 \times T \rceil$ represents the top 10\% of frames. The classification loss is then formulated as:
\begin{align}
\mathcal{L}_{\text{cls}} = -\frac{1}{N}\sum_{i=1}^{N}\big[&y_i\log(\sigma(s_{\text{video}}^{(i)})) \nonumber \\
&+ (1-y_i)\log(1-\sigma(s_{\text{video}}^{(i)}))\big].
\end{align}
where $N$ is the batch size, $y_i \in \{0,1\}$ is the video-level label, and $\sigma(\cdot)$ denotes the sigmoid function.
\subsubsection{Triplet Contrastive Loss}
To enhance the discriminative capability of learned representations, we introduce a triplet contrastive loss that enforces semantic consistency between model predictions and CLIP-generated pseudo-labels while maximizing the separation from normal samples:
\begin{align}
    \mathcal{L}_{\text{trip}} = \max(0, \|f_a - f_p\|_2^2 - \|f_a - f_n\|_2^2 + m).
\end{align}
where $f_a$ represents the anchor embedding from model predictions on anomalous segments, $f_p$ is the positive embedding derived from CLIP pseudo-labels on the same segments, $f_n$ is the negative embedding from normal segments, and $m$ is the margin parameter. This formulation encourages the model to learn representations consistent with CLIP's semantic understanding while maintaining clear boundaries between normal and anomalous patterns.
\subsubsection{Total Loss Function}
The complete loss function integrates the three complementary components through an uncertainty-weighted multi-task learning framework. We model the task-dependent homoscedastic uncertainty to automatically balance the relative contributions of each loss term without manual hyperparameter tuning.The total loss function is formulated as:
\begin{align}
\mathcal{L}_{\text{total}} &= \frac{1}{2\sigma_1^2}\mathcal{L}_{\text{PSE}} + \log(1+\sigma_1^2) \nonumber \\
&\quad + \frac{1}{2\sigma_2^2}\mathcal{L}_{\text{CLS}} + \log(1+\sigma_2^2) \nonumber \\
&\quad + \frac{1}{2\sigma_3^2}\mathcal{L}_{\text{TRI}} + \log(1+\sigma_3^2).
\end{align}

\section{Experiment and Analysis}
To validate the proposed DAMS method's superiority, it is compared with multiple state-of-the-art DAMS approaches on two large-scale datasets, namely, UCF-Crime and XD-Violence.

\subsection{Datasets}
\textbf{\emph{XD-Violence}} 
The XD-Violence dataset serves as a large-scale benchmark for weakly supervised video anomaly detection, specifically targeting violence detection, comprising 4,754 untrimmed videos (217 h) sourced from diverse real-world and online platforms. The dataset supports multimodal learning by incorporating both video and audio modalities. During training, only video-level weak labels are provided; however, the testing split includes frame-level start–end annotations to enable temporal localization. The dataset encompasses six categories of violent events: fighting, shooting, riot, abuse, explosion, and car accidents. Videos average approximately one minute in duration, recorded typically at 30 FPS with resolutions of approximately 320×240 pixels. The official data split contains 3,954 videos for training (60\% violent, 40\% non-violent) and 800 videos for testing (500 violent, 300 non-violent), resulting in a 83.2\%/16.8\% train/test distribution.

\textbf{\emph{UCF-Crime}}
The UCF-Crime dataset represents a widely adopted, large-scale surveillance benchmark comprising 1,900 untrimmed videos (128 h) acquired under diverse real-world conditions. The dataset includes 13 real-world anomalous event types (such as burglary, assault, vandalism, explosion, shoplifting, and robbery) alongside a normal activity class, with weak video-level labels exclusively indicating whether a video contains any anomaly, without providing temporal segment annotations. Videos typically last two to six minutes, recorded at approximately 320×240 resolution and 30 FPS. The dataset is partitioned into 1,610 videos for training (810 normal, 800 anomalous) and 290 videos for testing (150 normal, 140 anomalous). This partitioning underpins the conventional multiple-instance learning (MIL) evaluation framework.

\subsection{Evaluation Metrics}
For the XD-Violence dataset, frame-level Average Precision (AP) was primarily employed as a metric for anomaly detection performance, which directly reflects the algorithm's accuracy in identifying violent behaviours. For the UCF-Crime dataset, the algorithm's ability to distinguish between normal and abnormal videos was comprehensively evaluated using the frame-level Receiver Operating Characteristic (ROC) curve and its corresponding Area Under the Curve (AUC) metrics. The higher the AP and AUC values, the better the algorithmic performance and the greater the model robustness.

\subsection{Implementation Details}
All experiments in this study were conducted on workstations equipped with NVIDIA vGPU-48GB. Systematic hyperparameter optimization experiments were conducted using the Optuna framework to determine the optimal configuration for the DAMS model. For the AMTPN module, the scale parameters of temporal pyramid pooling were optimized and configured according to dataset-specific characteristics: the XD-Violence dataset and the UCF-Crime dataset utilized [1, 3, 9, 27] multi-scale configurations to capture temporal dependencies of anomalous events with varying durations.  The maximum number of training iterations is 5000, and validation evaluations are conducted every 100 steps. Automatic mixed precision training (AMP) was enabled for all experiments to enhance training efficiency and maintain numerical precision. During the data preprocessing stage, the training batch sizes for both the XD-Violence and UCF-Crime datasets were set to 30, with corresponding test batch sizes of 5 and 10, respectively. Data loading employed four parallel worker processes, and ten-crop data augmentation strategies were applied during the testing phase to improve model generalization performance and prediction stability.

\subsection{Comparison with State-of-the-art Methods}
\begin{table}[htbp]
\centering
\caption{{\fontfamily{ptm}\selectfont Comparative analysis of average precision (AP) performance with other benchmark methods under weakly supervised learning on XD-Violence dataset.}}
\label{tab4:tab}
\resizebox{\columnwidth}{!}{%
{\fontfamily{ptm}\selectfont
\begin{tabular}{l c c}
\hline
\textbf{Method} & \textbf{Feature} & \textbf{XD-Violence (AP \%)} \\
\hline
Wu et al. ~\cite{wu2020not}               & I3D-RGB     & 78.64 \\
RTSM ~\cite{tian2021weakly}               & I3D-RGB     & 77.81 \\
MSL  ~\cite{li2022self}                   & I3D-RGB     & 78.28 \\
NG-MIL  ~\cite{park2023normality}         & I3D-RGB     & 78.51 \\
CRFD  ~\cite{wu2021learning}              & I3D-RGB     & 75.90 \\
MGFN  ~\cite{chen2023mgfn}                & VS-RGB      & 79.19 \\
MGFN   ~\cite{chen2023mgfn}               & I3D-RGB     & 80.11 \\
CMSIL  ~\cite{qian2024clip}               & I3D-RGB     & 78.24 \\
ST-HTAM  ~\cite{paulraj2025transformer}   & Transformer & 78.06 \\ \hline
\textbf{DAMS (Ours)}                    & I3D-RGB     & \textbf{84.00} \\
\hline
\end{tabular}
}
}
\vspace{-0.5em}
\end{table}

\begin{table}[htbp]
\centering
\caption{{\fontfamily{ptm}\selectfont Comparative analysis of frame-level AUC (\%) with other benchmark methods under weakly supervised learning on the UCF-Crime dataset.}}
\label{tab5:tab}
\vspace{-0.3em}
\resizebox{\columnwidth}{!}{%
{\fontfamily{ptm}\selectfont
\begin{tabular}{l c c}
\hline
\textbf{Method} & \textbf{Feature} & \textbf{UCF-Crime (AUC \%)} \\
\hline
Sultani et al. ~\cite{sultani2018real}     & C3D-RGB     & 75.41 \\
TCN-IBL ~\cite{zhang2019temporal}           & C3D-RGB     & 78.66 \\
GCN  ~\cite{zhong2019graph}               & TSN-Flow    & 78.08 \\
GCN  ~\cite{zhong2019graph}                & C3D-RGB     & 81.08 \\
GCN  ~\cite{zhong2019graph}                & TSN-RGB     & 82.12 \\
MIST ~\cite{feng2021mist}              & C3D-RGB     & 81.40 \\
MIST ~\cite{feng2021mist}              & I3D-RGB     & 82.30 \\
RTSM ~\cite{tian2021weakly}              & C3D-RGB     & 83.28 \\
RTSM ~\cite{tian2021weakly}               & I3D-RGB     & 84.03 \\
BN-WVAD ~\cite{yi2022batch}            & I3D-RGB     & 84.29 \\
CRFD ~\cite{wu2021learning}               & I3D-RGB     & 89.89 \\
MSL ~\cite{li2022self}                & I3D-RGB     & 85.30 \\
NG-MIL ~\cite{park2023normality}            & I3D-RGB     & 85.63 \\
GLFE ~\cite{basak2024diffusion}              & I3D-RGB     & 86.12 \\
MGFN  ~\cite{chen2023mgfn}          & VS-RGB      & 86.67 \\
MGFN  ~\cite{chen2023mgfn}              & I3D-RGB     & 86.98 \\
CMSIL ~\cite{qian2024clip}             & I3D-RGB     & 77.07 \\
ST-HTAM  ~\cite{paulraj2025transformer}          & Transformer & 81.42 \\
TDS-Net ~\cite{hussain2024tds}           & I3D         & 84.50 \\
\hline
\textbf{DAMS (Ours)} & I3D-RGB & \textbf{94.67} \\
\hline
\end{tabular}
}
}
\end{table}

\subsubsection{Comparisons on XD-Violence}
Table~\ref{tab4:tab} presents a comprehensive quantitative comparison between the proposed DAMS and existing state-of-the-art methods on the XD-Violence dataset. The 5-crop I3D-RGB features provided by Wu et al.~\cite{wu2020not} were used as input for all experiments to ensure experimental fairness and reproducibility. As shown in the table, the proposed DAMS method achieves an average precision (AP) of 84.00\%, which is statistically significantly superior to existing methods. Specifically, compared with other methods utilizing I3D-RGB features, DAMS surpassed MGFN proposed by Chen et al.~\cite{chen2023mgfn}(79.19\%) by 4.81 percentage points. Compared with methods employing alternative feature extraction networks, DAMS demonstrates significant advantages, yielding performance 5.94 percentage points higher than ST-HTAM~\cite{paulraj2025transformer}, which utilizes Transformer features.  This result holds substantial significance as it demonstrates that the effectiveness of the DAMS architecture is independent of feature extractor complexity, but instead derives from its inherent architectural advantages. This significant improvement can be attributed to AMTPN's ability to effectively capture multiscale temporal dependencies, in conjunction with the synergistic effect of the CBAM attention mechanism for precise anomaly localization. 

\subsubsection{Comparisons on UCF-Crime}
To comprehensively evaluate the effectiveness of our method on the UCF-Crime dataset, we applied a ten-fold spatial sampling augmentation strategy to the extracted I3D RGB features, thereby generating high-dimensional 1024-dimensional feature representations for each input segment. In Table ~\ref{tab5:tab} through these feature enhancement techniques, the proposed model achieves an AUC@ROC of 94.67\%  on the UCF-Crime dataset, significantly outperforming all state-of-the-art methods reported in the current literature when evaluated under identical protocol conditions. These observed performance improvements can be attributed to the synergistic mechanism between the AMTPN and CBAM modules: AMTPN effectively captures anomalous patterns at different temporal granularities in sequential data through hierarchical multi-scale temporal pyramid pooling operations, while CBAM enables precise enhancement and localization of anomalous features across both channel and spatial dimensions. This dual complementary feature enhancement mechanism facilitates the model's capacity to address two temporal-scale anomalies in the UCF-Crime dataset: transient anomalous events and persistent anomalies with long-term evolution. When confronted with this complex evaluation environment, the proposed DAMS model demonstrates superior robustness and generalization capabilities, successfully differentiating low signal-to-noise ratio anomalous instances embedded within highly dynamic background activities, even under varying illumination conditions and camera viewpoints.

\subsection{Ablation Studies and Analysis} 

\begin{table}[htbp]
\centering
\caption{{\fontfamily{ptm}\selectfont Ablation study of CBAM and AMTPN on the XD-Violence and UCF-Crime datasets.}}
\label{tab6:tab}
\vspace{-0.3em}
\renewcommand{\arraystretch}{1.2}
\setlength{\tabcolsep}{5pt}
{\fontfamily{ptm}\selectfont
\begin{tabular}{ccc|c|c}
\hline
\textbf{Baseline} & \textbf{CBAM} & \textbf{AMTPN} & \makecell{\textbf{XD-Violence} \\ \textbf{(AP\%)}} & \makecell{\textbf{UCF-Crime} \\ \textbf{(AUC\%)}} \\
\hline
\checkmark &              &              & 77.07 & 78.24 \\
\checkmark & \checkmark   &              & 80.60 & 85.29 \\
\checkmark &              & \checkmark   & 83.01 & 87.45 \\
\hline
\checkmark & \checkmark   & \checkmark   & \textbf{84.00} & \textbf{94.67} \\
\hline
\end{tabular}
}
\end{table}

\begin{table}[htbp]
\centering
\caption{{\fontfamily{ptm}\selectfont Ablation study of AMTPN internal components on the XD-Violence and UCF-Crime datasets.}}
\label{tab7:tab}
\vspace{-0.3em}
\renewcommand{\arraystretch}{1.2} % 行高调整
\setlength{\tabcolsep}{5pt}       % 列间距调整
{\fontfamily{ptm}\selectfont
\begin{tabular}{cccc|c|c}
\hline
\textbf{Baseline} & \textbf{AFF} & \textbf{TCE} & \textbf{TPP} & \makecell{\textbf{XD-Violence} \\ \textbf{(AP\%)}} & \makecell{\textbf{UCF-Crime} \\ \textbf{(AUC\%)}} \\
\hline
\checkmark &        &        &        & 77.07 & 78.24 \\
\checkmark &        & \checkmark & \checkmark & 79.35 & 82.18 \\
\checkmark & \checkmark &        & \checkmark & 82.77 & 91.15 \\
\checkmark & \checkmark & \checkmark &        & 82.02 & 89.43 \\
\hline
\checkmark & \checkmark & \checkmark & \checkmark & \textbf{84.00} & \textbf{94.67} \\
\hline
\end{tabular}
}
\end{table}

\begin{table}[htbp]
\centering
\caption{{\fontfamily{ptm}\selectfont Ablation study of CBAM components (SA: Spatial Attention, CA: Channel Attention) on XD-Violence and UCF-Crime datasets.}}
\label{tab8:tab}
\vspace{-0.3em}
\renewcommand{\arraystretch}{1.2}
\setlength{\tabcolsep}{8pt}
{\fontfamily{ptm}\selectfont
\begin{tabular}{c c c|c|c}
\hline
\textbf{Baseline} & \textbf{SA} & \textbf{CA} & \makecell{\textbf{XD-Violence} \\ \textbf{(AP\%)}} & \makecell{\textbf{UCF-Crime} \\ \textbf{(AUC\%)}} \\
\hline
\checkmark &            &            & 77.07 & 78.24 \\
\checkmark & \checkmark &            & 83.63 & 88.73 \\  
\checkmark &            & \checkmark & 82.88 & 90.73 \\ 
\hline
\checkmark & \checkmark & \checkmark & \textbf{84.00} & \textbf{94.67} \\
\hline
\end{tabular}
}
\end{table}

\begin{table}[htbp]
\centering
\caption{{\fontfamily{ptm}\selectfont Ablation study of loss components on the XD-Violence and UCF-Crime datasets.}}
\label{tab9:tab}
\vspace{-0.3em}
\renewcommand{\arraystretch}{1.2}
\setlength{\tabcolsep}{10pt}
{\fontfamily{ptm}\selectfont
\begin{tabular}{ccc|c|c}
\hline
\textbf{$\ell_{\text{cls}}$} & \textbf{$\ell_{\text{trip}}$} & \textbf{$\ell_{\text{pse}}$} & \makecell{\textbf{XD-Violence} \\ \textbf{(AP\%)}} & \makecell{\textbf{UCF-Crime} \\ \textbf{(AUC\%)}} \\
\hline
\checkmark & \checkmark &             & 79.00 & 71.83 \\
\checkmark &            & \checkmark  & 80.67 & 91.41 \\
\hline
\checkmark & \checkmark & \checkmark  & \textbf{84.00} & \textbf{94.67} \\
\hline
\end{tabular}
}
\end{table}

We performed comprehensive ablation experiments to understand further the contributions of the components of our proposed DAMS architecture. These experiments aimed to separate and evaluate the impact of each major architectural component. Throughout these experiments, all hyperparameters (including network architecture parameters, training strategy parameters, and loss function parameters) were maintained constant to ensure consistency and reliability of the results. To assess the complete model's effectiveness, we systematically constructed and evaluated various baseline models by selectively removing or modifying specific architectural components. The complete DAMS model serves as the reference standard for comparison. The model integrates two core components: an adaptive multi-scale time pyramid network (AMTPN) and a convolutional block attention mechanism (CBAM). The experimental results and corresponding analyses are presented below:

\subsubsection{Role of AMTPN}

\begin{figure}[htbp]
    \centering
    \includegraphics[width=\linewidth]{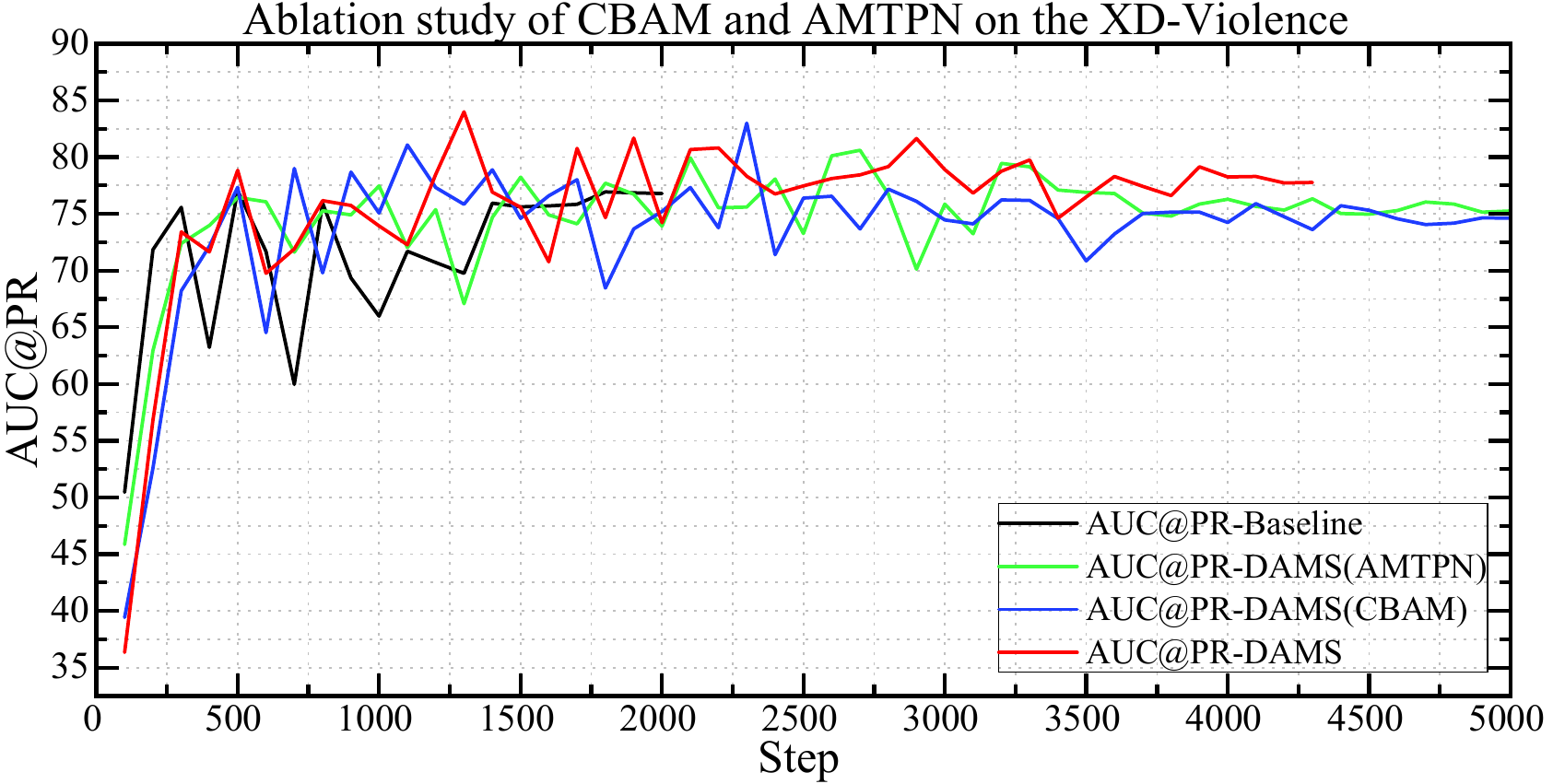}
    \caption{{\fontfamily{ptm}\selectfont Ablation study of CBAM and AMTPN on the XD-Violence dataset.}}
    \label{fig:fig6}
    \vspace{-0.2cm}
\end{figure}

\begin{figure}[htbp]
    \centering
    \includegraphics[width=\linewidth]{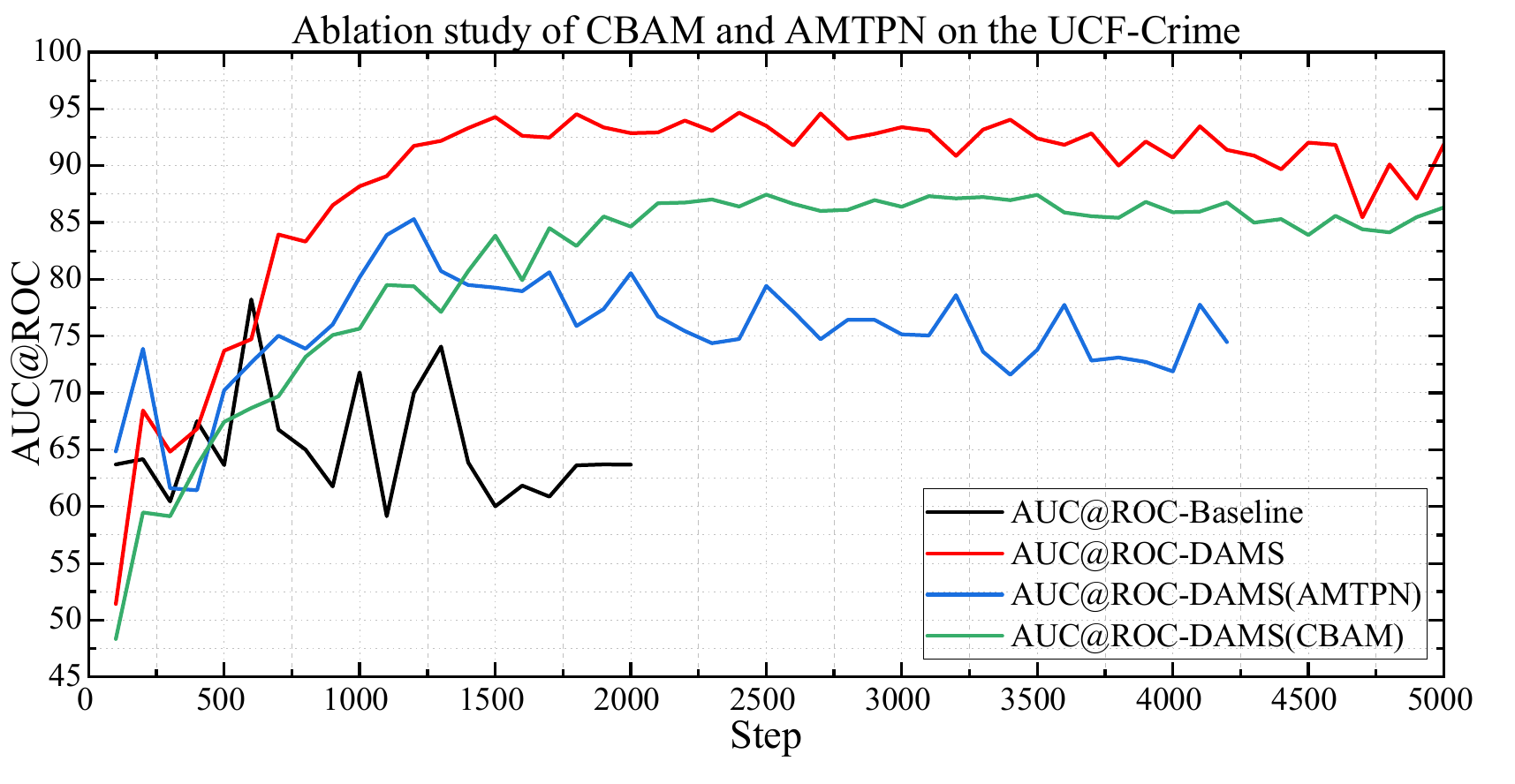}
    \caption{{\fontfamily{ptm}\selectfont Ablation study of CBAM and AMTPN on the UCF-Crime dataset.}}
    \label{fig:fig7}
    \vspace{-0.2cm}
\end{figure}

As a core innovative contribution, the Adaptive Multiscale Temporal Pyramid Network (AMTPN) is designed to capture multiscale temporal dependencies in video sequences. Controlled ablation experiments compared the full model against variants with specific AMTPN components removed. As shown in Table~\ref{tab6:tab} and Figure ~\ref{fig:fig6},~\ref{fig:fig7}, AMTPN substantially enhances anomaly detection performance, achieving improvements of 17.6\% in AUC@ROC on UCF-Crime and 10.80\% in AUC@PR on XD-Violence. The contributions of three key components—Adaptive Feature Fusion (AFF), Temporal Context Enhancement (TCE), and Temporal Pyramid Pooling (TPP) are detailed in Table~\ref{tab7:tab}. The AFF module had the most significant impact, with its removal causing performance drops of 12.49\%  on UCF-Crime and 4.65\% on XD-Violence. The TPP module contributed notable gains of 5.24\% and 1.98\%, respectively, while the TCE module delivered increases of 3.52\% and 1.23\%. These components exhibit significant synergistic effects, demonstrating complementarity in capturing multiscale temporal information.

\subsubsection{Influence of CBAM}
The convolutional block attention module (CBAM) serves as a feature enhancement mechanism, selectively enhancing discriminant features through dual-stream processing with channel attention (CA) and spatial attention (SA). Controlled experiments systematically evaluated its internal mechanisms, analyzing performance differences between models using only CA, SA, or the fully integrated CBAM architecture.
As illustrated in Table~\ref{tab8:tab} and Figure ~\ref{fig:fig6},~\ref{fig:fig7}, the complete CBAM framework achieved detection performance of 94.67\% AUC@ROC and 84.00\% AUC@PR on UCF-Crime and XD-Violence datasets, respectively, statistically significantly outperforming models using only CA (UCF-Crime: 90.73\%, XD-Violence: 82.88\%) or only SA (UCF-Crime: 88.73\%, XD-Violence: 83.63\%). These results verify the dual attention mechanism's synergistic effect, demonstrating that channel and spatial domain feature selection complementarity is theoretically and practically significant for anomaly detection. Notably, CBAM's performance gain exceeds the expected theoretical value of simple CA and SA superposition, quantitatively confirming nonlinear synergistic effects between the two mechanisms.

\subsubsection{The effectiveness of different loss terms}
To systematically evaluate the impact of the three loss terms, this section presents quantitative results for different loss function combinations in Table~\ref{tab9:tab}. The model employs $\mathcal{L}_{cls}$ as the primary supervision signal, utilizing a multi-objective optimization framework to guide training. Empirical analysis demonstrates that triplet loss $\mathcal{L}{trip}$ exhibits superior performance in enhancing discriminative capacity, yielding gains of 22.84\% and 5.00\% on UCF-Crime and XD-Violence datasets respectively.  This corroborates the theoretical significance of contrastive learning in feature representation. $\mathcal{L}{trip}$ enhances discriminability by maximizing Euclidean distance between anomalous and normal samples in feature embedding space, particularly for anomalous events with high visual similarity but semantic divergence. The pseudo-label loss $\mathcal{L}_{pse}$ generated using CLIP yields improvements of 3.26\% and 3.33\% on the respective datasets. Leveraging cross-modal knowledge transfer addresses the challenge of absent frame-level annotations under weak supervision. Although pseudo-label loss's independent contribution is modest, its synergistic effect with contrastive loss demonstrates that integrating cross-modal knowledge and contrastive learning produces super-linear performance improvements. 

\subsubsection{Temporal Anomaly Score Comparison Visualization}
\begin{figure*}[htbp]
    \centering
    \includegraphics[width=\linewidth]{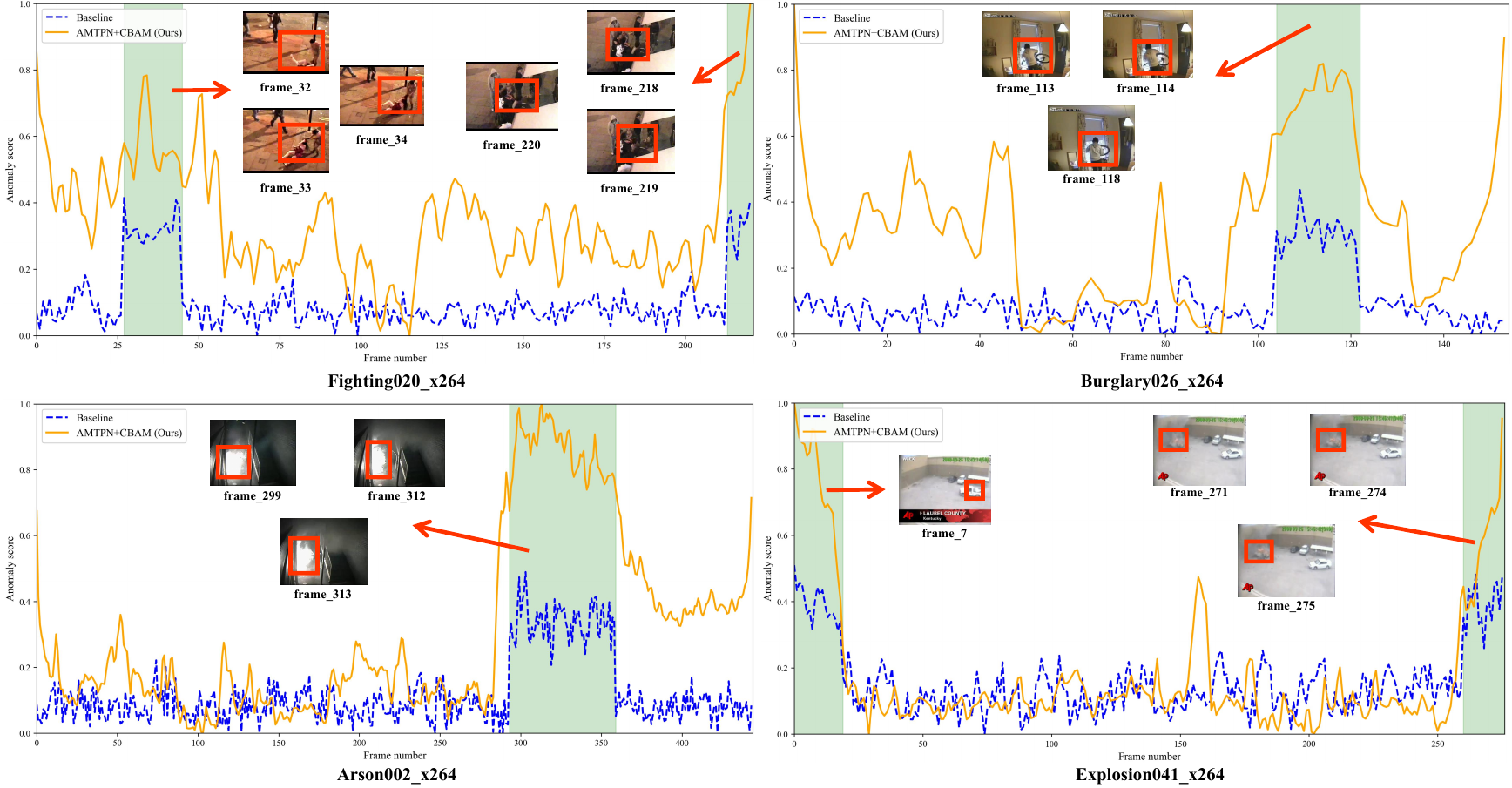}
    \caption{{\fontfamily{ptm}\selectfont Temporal anomaly score comparison visualization.}}
    \label{fig3.pdf}
    %\vspace{-0.2cm}
\end{figure*}
We present a qualitative comparison of anomaly scores in Figure~\ref{fig3.pdf}, where the blue dashed curve represents the anomaly score of the baseline model, the yellow solid curve represents the anomaly score of our proposed model, and the light green shaded regions indicate the ground truth anomalous segments. The results demonstrate that AMSTAD effectively identifies different categories of anomalous regions across multiple scenarios, while maintaining low anomaly scores for normal frames.

\subsubsection{Visualization of spatiotemporal dimensions of feature representation evolution}
\begin{figure}[htbp]
    \centering
    \includegraphics[width=\linewidth]{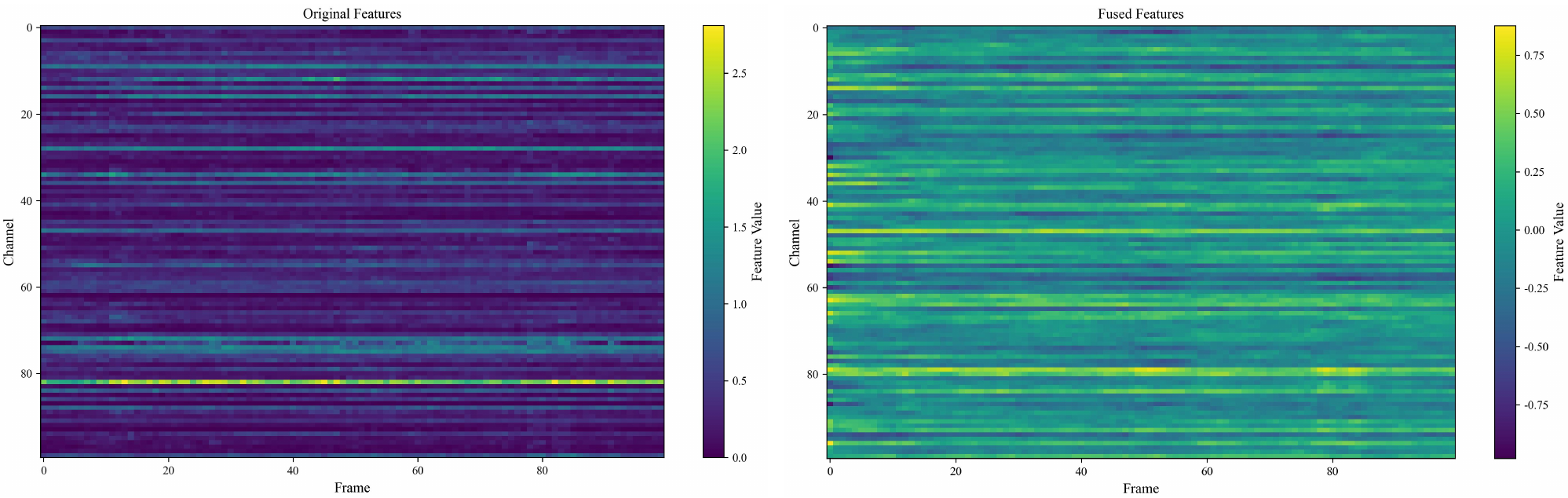}
    \caption{{\fontfamily{ptm}\selectfont Visualization of spatiotemporal dimensions of feature representation evolution.}}
    \label{fig4.pdf}
\end{figure}
Figure~\ref{fig4.pdf} illustrates how our proposed method evolves feature representations for video anomaly detection. The left panel displays the original extracted feature map, while the right panel shows the optimized feature representation following adaptive multi-scale spatiotemporal information fusion. Comparative analysis reveals that the fused features, when contrasted with the original features, demonstrate a transformation from single positive value representation to a centralized distribution, indicating a more stable feature normalization in a statistical sense. Additionally, the activation mode transitions from a single channel to a multi-channel coordinated expression, thereby significantly enhancing the features' discriminative capability and representational efficiency. Furthermore, the temporal coherence is significantly enhanced, resulting in smoother feature transitions between adjacent frames, which reflects the effective integration of long-range temporal dependencies through the attention mechanism. Finally, the information distribution becomes more balanced, reducing feature redundancy and optimizing information entropy. This optimization of hierarchical feature representation plays a key role in improving the accuracy and robustness of video anomaly detection under weakly supervised conditions.

\subsubsection{Visualization of Multi-scale Temporal Pyramid Pooling Features}
\begin{figure}[htbp]
    \centering
    \includegraphics[width=\linewidth]{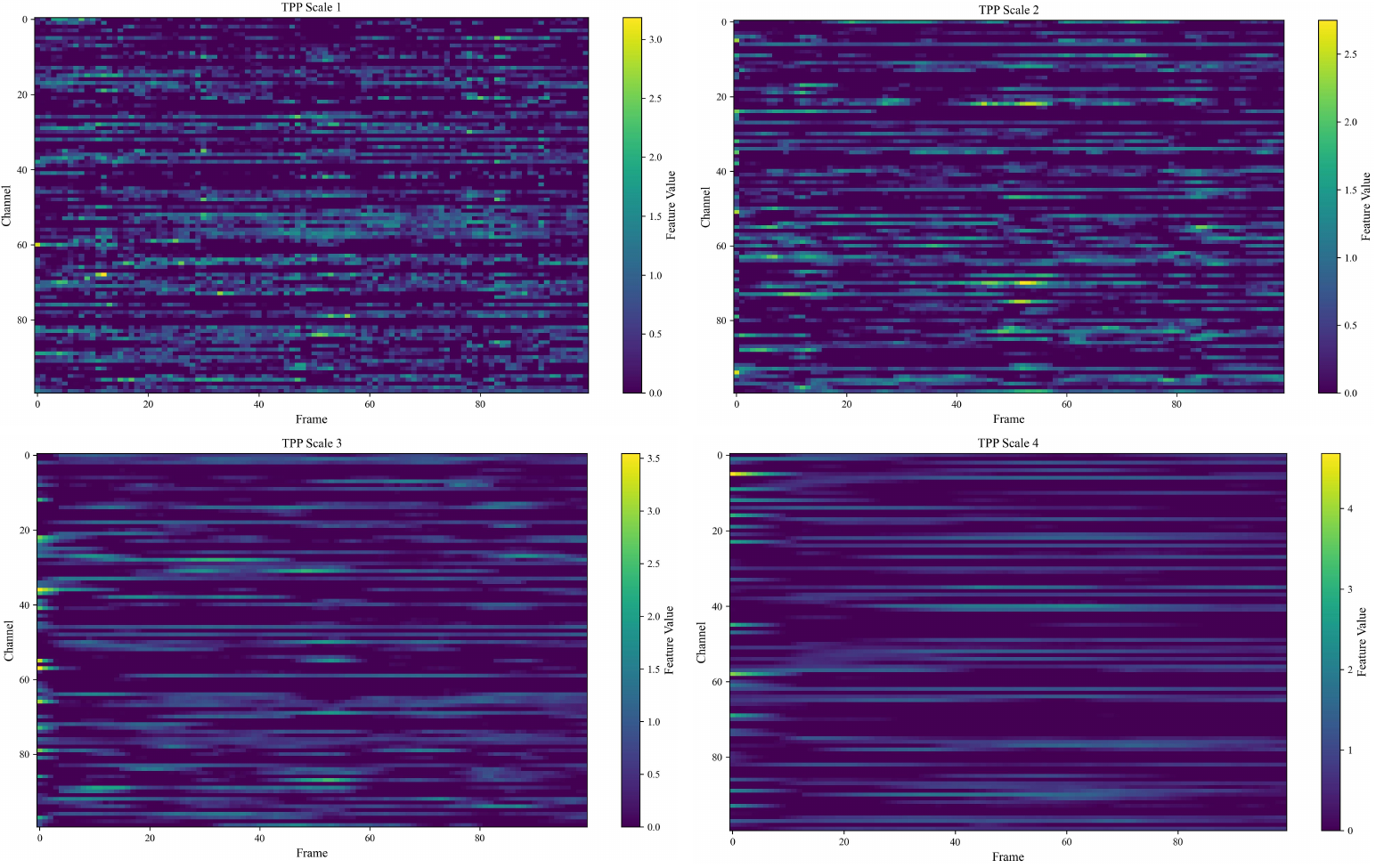}
    \caption{{\fontfamily{ptm}\selectfont Visualization of multi-scale temporal pyramid pooling features.}}
    \label{fig:fig5}
\end{figure}
Figure~\ref{fig:fig5} illustrates the feature activation distributions across four different temporal scales (Scale 1-4), with the horizontal axis representing the temporal dimension and the vertical axis representing the feature channel dimension. Scale 1 primarily captures local short-term features, Scale 2 facilitates the formation of preliminary temporal coherence, Scale 3 enhances mid-term temporal correlation, while Scale 4 extracts long-term semantic information. Our experimental results demonstrate that as the temporal scale increases, feature representation progressively transitions from high-frequency details to low-frequency semantic representations, with high-level features exhibiting significant selective activation patterns for anomalies. This hierarchical feature extraction mechanism addresses the inherent trade-off between temporal accuracy and long-term dependencies. It provides a robust foundation for feature representation in video anomaly detection under weakly supervised conditions.

\section{Conclusion} 
In this work, we propose the Adaptive Multiscale Spatial-Temporal Attention Framework (DAMS), a dual-branch architecture for weakly supervised video anomaly detection that effectively addresses the challenges of multiscale spatial-temporal modelling and semantic representation. The framework integrates an adaptive multiscale temporal pyramid network (AMTPN) with the main path's convolutional block attention module (CBAM). It utilises pseudo-labels generated by CLIP to provide semantic guidance. Extensive experiments conducted on the UCF-Crime and XD-Violence datasets demonstrate significant performance improvements compared to existing state-of-the-art methods. The main contributions include: (1) proposing a novel temporal pyramid architecture with adaptive feature fusion that captures multiscale temporal dependencies; (2) integrating channel and spatial attention mechanisms to enhance feature discriminative capabilities; and (3) effectively leveraging vision-language pre-training through pseudo-label generation. The orthogonal complementarity between spatiotemporal modelling and semantic understanding proves critical for robust anomaly detection. Despite this progress, several limitations remain. First, relying on pre-computed CLIP pseudo-labels may introduce domain bias and limit adaptability to new anomaly types. Second, the computational overhead limits real-time deployment. Future work will explore end-to-end trainable architectures and more efficient attention mechanisms to address these limitations while preserving detection accuracy. Furthermore, we plan to investigate semi-supervised learning methods that better utilise unlabeled data and explore multimodal fusion strategies beyond CLIP to incorporate additional semantic modalities such as audio and motion cues.

% Numbered list
% Use the style of numbering in square brackets.
% If nothing is used, default style will be taken.
%\begin{enumerate}[a)]
%\item 
%\item 
%\item 
%\end{enumerate}  

% Unnumbered list
%\begin{itemize}
%\item 
%\item 
%\item 
%\end{itemize}  

% Description list
%\begin{description}
%\item[]
%\item[] 
%\item[] 
%\end{description}  

% Uncomment and use as the case may be
%\begin{theorem} 
%\end{theorem}

% Uncomment and use as the case may be
%\begin{lemma} 
%\end{lemma}

%% The Appendices part is started with the command \appendix;
%% appendix sections are then done as normal sections
%% \appendix

%% Loading bibliography style file
%\bibliographystyle{model1-num-names}
\bibliographystyle{unsrt}
\bibliography{ref}  % references.bib 文件

% Loading bibliography database

% Biography
%\bio{}
% Here goes the biography details.
%\endbio

%\bio{pic1}
% Here goes the biography details.
%\endbio

\end{document}